%% file: main.tex
\PassOptionsToPackage{table}{xcolor}
\documentclass[10pt,twocolumn,letterpaper]{article}
\usepackage[pagenumbers]{cvpr}      

\input{preamble}

\definecolor{cvprblue}{rgb}{0.21,0.49,0.74}
\usepackage[breaklinks,colorlinks,allcolors=cvprblue,plainpages=false]{hyperref}

\def\confName{CVPR}
\def\confYear{2027}

\title{SPIRAL-360: Stratified Panoramic Instruction Replay and Angular Learning for Visual Grounding}
\author{
Jingtao Xu\textsuperscript{1}\quad
Zizhuo Lin\textsuperscript{1}\quad
Jianwen Sun\textsuperscript{2}\quad
Yi Yang\textsuperscript{1}\quad
Yawei Luo\textsuperscript{1}\thanks{Corresponding author.}\\[0.5em]
\textsuperscript{1}Zhejiang University\qquad
\textsuperscript{2}Central China Normal University
}
\hypersetup{pdftitle={SPIRAL-360: Stratified Panoramic Instruction Replay and Angular Learning for Visual Grounding},pdfauthor={Jingtao Xu, Zizhuo Lin, Jianwen Sun, Yi Yang, Yawei Luo}}

\begin{document}
\maketitle
\input{sec/0_abstract}
\input{sec/1_teaser}
\input{paper_body}
\FloatBarrier
\input{Section/ai_use_statement}

{\small
\bibliographystyle{ieeenat_fullname}
\bibliography{main}
}

\end{document}

%% file: preamble.tex
\usepackage{microtype}
\usepackage{multirow}
\usepackage{placeins}
\usepackage{stfloats} 
\usepackage[skins,breakable]{tcolorbox}
\input{math_commands.tex}

\newtcolorbox{promptbox}[1][]{
    enhanced, breakable,
    colback=gray!5!white, colframe=gray!75!black,
    fonttitle=\bfseries\sffamily, title=#1,
    boxrule=0.5pt, arc=4pt,
    left=6pt, right=6pt, top=6pt, bottom=6pt
}

%% file: math_commands.tex
\usepackage{amsmath,amsfonts,bm}

\def\eqref#1{equation~\ref{#1}}

\def\1{\bm{1}}

\DeclareMathAlphabet{\mathsfit}{\encodingdefault}{\sfdefault}{m}{sl}
\SetMathAlphabet{\mathsfit}{bold}{\encodingdefault}{\sfdefault}{bx}{n}



%% file: sec/0_abstract.tex
\begin{abstract}
Localizing targets in $360^\circ$ environments requires combining global context with fine-grained evidence despite projection distortion and horizontal wrap-around. Uneven directional coverage further limits learning, particularly for polar targets. We present SPIRAL-360, a geometry-aware framework with separately trained single-turn and multi-turn models for panoramic localization. Spherically Stratified Replay (SSR) broadens directional coverage through stratified sampling and consistent transformations of panoramas, target annotations, and tool trajectories. TapSphere incorporates spherical distance priors into panoramic visual attention, while Spherical Direction Regularization (SDR) complements coordinate-token prediction with auxiliary direction supervision during training. The single-turn model predicts directly from the panorama, while the multi-turn model inspects tool-generated perspective views. We construct SPIRAL-360 Search, a curated dataset of trajectories interleaving reasoning, projection actions, and visual observations. Both models undergo full-parameter supervised fine-tuning followed by policy optimization. Experiments demonstrate state-of-the-art localization on S360 and competitive cross-dataset transfer, including H$^*$Bench. Systematic ablations support the effectiveness of this geometric learning design.
\end{abstract}

%% file: sec/1_teaser.tex
\begin{figure*}[!t]
  \centering
  \includegraphics[width=\textwidth]{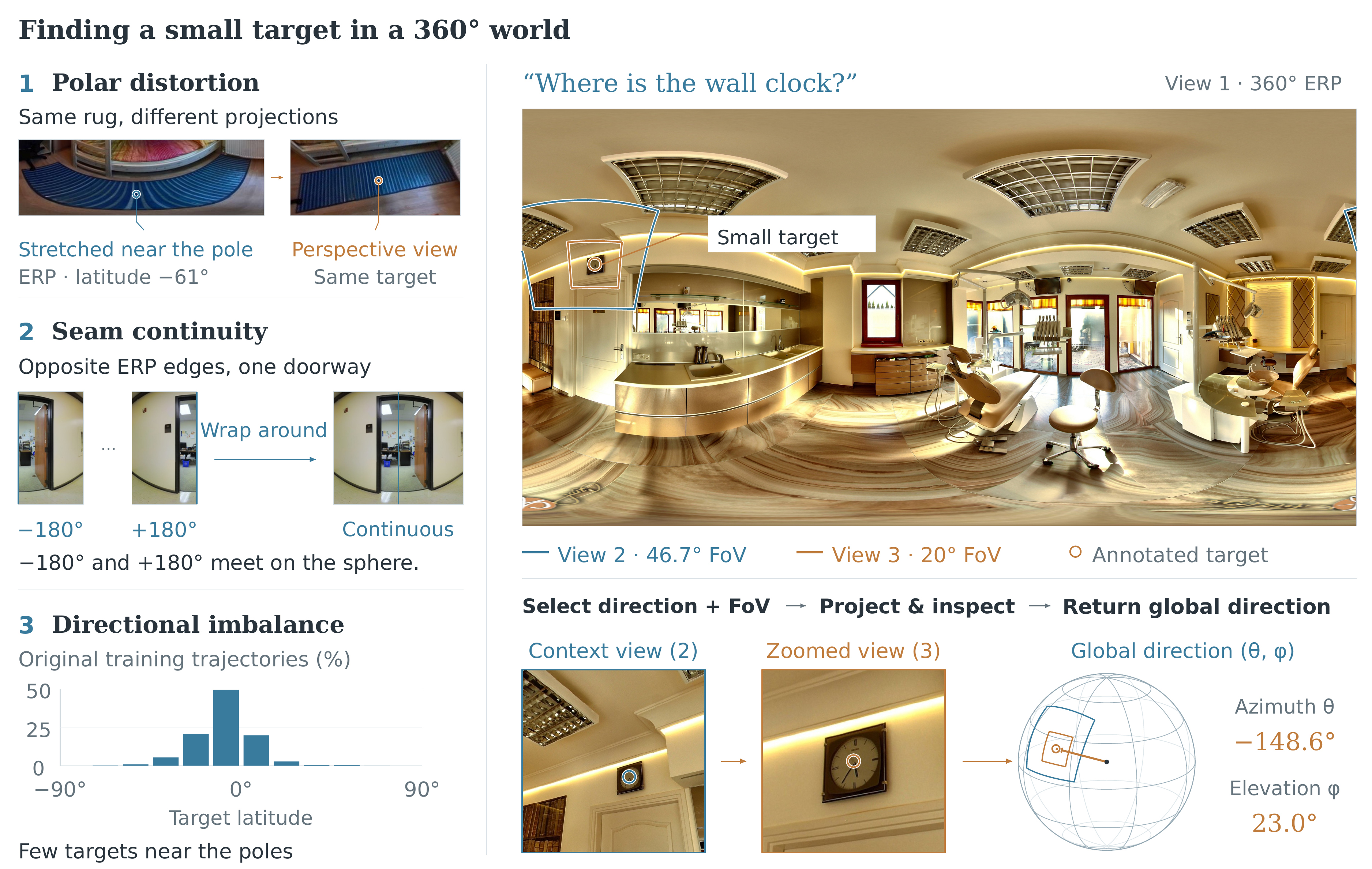}
  \caption{\textbf{Panoramic search and its geometric challenges.} Left: polar distortion, cyclic seam continuity, and target-latitude frequencies. Right: perspective inspection and global direction output from one fixed panorama. View~1 is the initial ERP; displayed Views~2/3 show recorded tool observations. Outlines show their actual fields of view.}
  \label{fig:teaser}
\end{figure*}

%% file: paper_body.tex
\section{Introduction}

Language-guided panoramic search locates a described target anywhere in a $360^\circ$ scene and reports its direction in global spherical coordinates (Figure~\ref{fig:teaser}). Existing approaches range from local-view search, as in Hstar, to full-panorama grounding, as in PanoWorld~\cite{Yu_2025_Thinking360,Wang_2026_PanoWorld}. A panorama provides global context for direct prediction, while perspective tools enable closer inspection of local evidence. These approaches offer complementary trade-offs between localization accuracy and inference efficiency.

Panoramic search poses several geometric challenges. Equirectangular projection (ERP) separates neighboring regions at the horizontal seam and distorts object appearance and image-space distances near the poles. Uneven training coverage further underrepresents polar targets. Finally, coordinate-token likelihood does not explicitly measure angular proximity. These challenges motivate geometric learning across data construction, visual attention, and direction supervision.

We introduce \textbf{SPIRAL-360}, a geometry-aware framework with separately trained single-turn and multi-turn models for panoramic search (Figure~\ref{fig:pipeline}). The single-turn model predicts a global direction directly from the panorama. The multi-turn model interleaves reasoning with visual tool use: it selects a viewing direction and field of view (FoV), inspects the returned perspective image, and incorporates the evidence into subsequent reasoning before answering.

Both models share three components. Spherically Stratified Replay (SSR) broadens directional coverage through stratified sampling and synchronized transformations of panoramas, target directions, camera actions, and coordinate-dependent text. Re-rendering tool observations preserves trajectory consistency. TapSphere introduces spherical distance priors into attention over panoramic patches while retaining native positional processing for perspective images. Spherical Direction Regularization (SDR) complements coordinate-token supervision with a training-only direction head that aligns hidden representations with spherical targets. Final coordinates remain language-generated.

We construct SPIRAL-360 Search (S360-Search), linking panoramas and queries to target directions and trajectories of reasoning, projection actions, and observations. The training pool contains 24,698 trajectories including SSR replay, while direction-balanced S360 evaluation includes controlled reorientations. We derive direct-answer examples for single-turn training and retain full trajectories for multi-turn training. Both models undergo supervised fine-tuning followed by policy optimization.

SPIRAL-360 achieves state-of-the-art S360 localization among the evaluated systems. Single-turn inference avoids tool calls and uses fewer visual and generated tokens, while multi-turn inference achieves a higher S360 hit rate. These results support visual tool reasoning and characterize its accuracy--efficiency trade-off. Transfer evaluation on APRS256 and H$^*$Bench demonstrates competitive object localization beyond S360, and controlled ablations support SSR, TapSphere, and SDR under the evaluated settings.

Our contributions are threefold:
\begin{itemize}
    \item We introduce SPIRAL-360, combining shared geometric learning with separately trained models for efficient direct prediction and multi-turn visual tool reasoning.
    \item We construct S360-Search, a panoramic grounding corpus with multi-turn training trajectories and direction-balanced evaluation for comparing both models.
    \item We demonstrate state-of-the-art S360 performance among evaluated systems and competitive object localization on APRS256 and H$^*$Bench, supported by component ablations and analysis of the accuracy--efficiency trade-off.
\end{itemize}

\input{Section/2_related_work}

\section{Method}
\label{sec:method}
\begin{figure*}[!t]
  \centering
  \includegraphics[width=\textwidth]{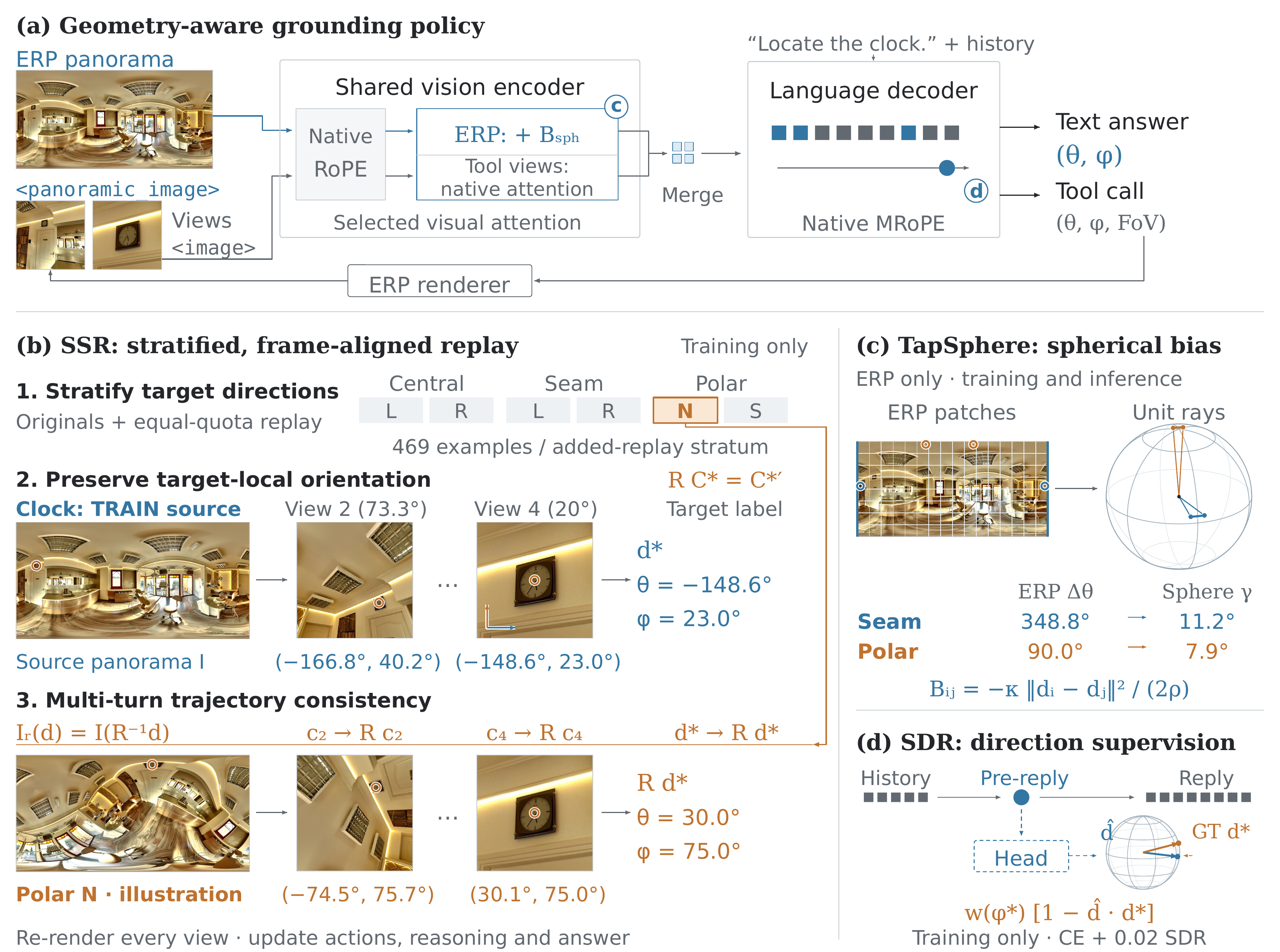}
  \caption{\textbf{SPIRAL-360 training and interaction.} \textbf{(a)} Shared visual encoding and language decoding, with projected views appended for multi-turn inference. \textbf{(b)} SSR combines directional stratification, target-local frame alignment, and trajectory reconstruction. \textbf{(c)} TapSphere adds an ERP-only attention bias. \textbf{(d)} SDR supervises hidden states before each reply during training. Angles above views denote FoV.}
  \label{fig:pipeline}
\end{figure*}

Figure~\ref{fig:pipeline} summarizes the shared training framework. SSR expands directional coverage through trajectory reconstruction, TapSphere adds a spherical prior to ERP attention, and SDR supervises direction representations before each assistant reply. We use these components to train separate single-turn and multi-turn models.

\subsection{Panoramic grounding and tool interaction}
\label{subsec:problem_formulation}
\label{subsec:exploration_framework}
Given a panorama $I$ and text query $q$, the model predicts the target's global azimuth and elevation $(\widehat\theta,\widehat\phi)$. Angles $\theta\in[-\pi,\pi)$ and $\phi\in[-\pi/2,\pi/2]$ define the unit ray
$\mathbf d(\theta,\phi)=(\cos\phi\cos\theta,\cos\phi\sin\theta,\sin\phi)$.
All views share a fixed optical center. A tool call specifies a viewing direction $(\theta_t,\phi_t)$ and FoV $f_t$. The renderer projects $I$ into a perspective view and appends it with a turn prompt to the dialogue history; the tool call retains the camera parameters. The ERP and tool views share a vision encoder; online rendering uses no target annotations.

The multi-turn response format comprises reasoning followed by a tool call or terminal answer. The single-turn model answers from the initial panorama alone. Both report global spherical coordinates. Invalid answers and tool-budget exhaustion count as failures. For predicted and target rays, angular error and the annotation-dependent hit criterion are
\begin{equation}
\begin{aligned}
d_{\rm gc}&=\arccos\!\left(\operatorname{clip}_{[-1,1]}
       (\widehat{\mathbf d}^{\top}\mathbf d^*)\right),\\
\mathrm{Hit}_{\rm tol}&=\mathbf 1[d_{\rm gc}\leq\delta_{\rm bbox}/2],
\end{aligned}
\label{eq:bfov_eval}
\end{equation}
where $\delta_{\rm bbox}$ is the annotated angular diagonal. We also report fixed-$5^\circ$ accuracy, valid-answer rate, and inference cost. Invalid outputs contribute $180^\circ$ to mean error; the last tool-view direction is not treated as an answer. Equations use radians; reported errors and tool arguments use degrees.

\input{Section/3_tapsphere}
\input{Section/3_geometry_aware_training}
\input{Section/3_supervised_modes}

\section{Dataset}
\label{sec:dataset}
\input{Section/current_main_results_table}
\paragraph{Sources and scale.}
\textbf{SPIRAL-360 Search} (\textbf{S360-Search}) combines Matterport3D~\cite{Matterport3D}, 2D-3D-S~\cite{armeni2017joint}, Kuula~\cite{kuula}, and curated PanoInfinigen~\cite{bozic2026pager} examples. Its 5,471 target records span 4,882 source panoramas, not necessarily distinct physical objects or scenes. Four yaw variants yield 21,884 trajectories; 2,814 SSR replays expand training to 24,698 examples.

\paragraph{Annotation and reference trajectories.}
\label{subsec:trajectory_synthesis}
\label{subsec:trajectory_formatting}
Geometric checks and GroundingDINO~\cite{liu2023grounding} verify proposals in projected views. GPT-4o-mini~\cite{openai2024gpt4omini} assists real-image descriptions and uniqueness checks; Qwen3-VL~\cite{Qwen3-VL} screens expansion examples. Reference trajectories combine annotation-guided cameras, rendered observations and geometric reasoning templates. The pool contains 124,103 assistant replies, including 99,405 tool requests, supporting full-dialogue and direct-answer supervision (Section~\ref{subsec:training}). Annotation details appear in the supplement.

\paragraph{Splits and directional coverage.}
Development uses 319 PanoInfinigen questions; testing uses 360 real-image and 353 PanoInfinigen questions. Each has one fixed orientation. Jointly reorienting 111 development and 235 test views with their annotations balances directions; other views retain source poses. The resulting 713-question test split is \textbf{S360}. Before reorientation, no development target and only one test target had $|\phi|\geq60^\circ$. These splits test controlled directional shifts, without establishing natural-polar generalization. Panorama identifiers and available source hashes screen overlaps; missing room/camera mappings prevent verified physical-scene isolation.

\section{Experiments}
\label{sec:experiments}

\subsection{Benchmarks and protocols}
\label{subsec:experiment_setup}
\paragraph{Evaluation benchmarks.}
S360 assesses localization under controlled directional shifts using the 713-question, direction-balanced test split of S360-Search (Section~\ref{sec:dataset}). H$^*$Bench~\cite{Yu_2025_Thinking360} contains 600 object-search and 400 path-search questions. We evaluate each once from full ERP; the original protocol uses four restricted-view starting headings. APRS256~\cite{tang2026seek} adapts TEST-UNIQ to full-ERP point localization and includes 11 polar components; APRS128 provides a later same-source check. Both retain source poses and omit official active-search segmentation. APRS inputs were fixed before scoring, although APRS256 was subsequently used during development. The supplement details selection and coverage.

\paragraph{Training protocol.}
Full-parameter SFT uses Qwen3-VL-8B-Instruct, panoramic tokens, seed 42, eight A100-80GB GPUs, and a global batch size of eight. One epoch comprises 24,698 SSR-augmented examples and 3,088 updates, or 21,884 original examples and 2,736 updates. Native retains native positional encoding; geometry controls share SSR and CE+$0.02\mathcal L_{\rm SDR}$. GRPO continues each mode's TapSphere SFT through language LoRA for 280/100 single-/multi-turn updates.

\paragraph{Inference and metrics.}
By default, we use 512 ERP visual tokens, greedy decoding, and at most 512 generated tokens per reply. Multi-turn inference allows six tool views at 256 visual tokens each; public baselines retain their own protocols. Hit measures the percentage of questions with angular error within half the annotated angular-box diagonal (Eq.~\ref{eq:bfov_eval}); F5 uses a fixed $5^\circ$ threshold. GCD averages great-circle error in degrees over all questions. Invalid answers fail Hit/F5 and receive $180^\circ$ error. These metrics capture annotation-scaled success, fixed-angle precision, and error magnitude; GCD also reflects answer completion. H$^*$Bench uses native task-specific success rates. Paired intervals use source-group bootstrap.

\subsection{System comparison and post-training}
\label{subsec:main_results}
Combining SSR, TapSphere, SDR, and GRPO, SPIRAL-360 achieves state-of-the-art S360 results among the compared systems (Table~\ref{tab:current_main_results}). Single-/multi-turn Hit reaches 89.62/91.16\%, compared with the original Qwen3-VL-8B's 5.61\%, and exceeds all evaluated proprietary models (best: 71.53\%) and open-source baselines (best: 82.75\%, \mbox{GroundingPI}). GCD is $7.83/8.48^\circ$, also below every compared baseline. Gains over \mbox{GroundingPI} concentrate in Up/Down. Multi-turn inference achieves higher Hit, while single-turn inference has lower error and token cost.

\input{Section/current_external_results_table}
On H$^*$Bench (Table~\ref{tab:current_hstar_results}), SPIRAL-360 achieves state-of-the-art zero-shot HOS performance among the compared methods. Single-/multi-turn success reaches 68.33/66.17\%, both above PanoWorld's 61.8\%, without H$^*$-specific training. HPS success is 25.25/24.25\%; path search lies outside our training objective. On jointly valid questions, multi-turn lowers mean HOS error from $27.60^\circ$ to $21.03^\circ$ and HPS yaw error from $62.70^\circ$ to $51.98^\circ$. However, its valid-answer rates are lower (81.33/62.50\% versus 98.67/90.75\%). Requests for additional observations beyond the six-view limit account for 91.22\% of its invalid episodes, limiting overall success despite finer localization on jointly valid answers.

GRPO improves S360 F5 and all-question GCD in both modes. Single-turn F5 rises from 553 to 594 successes ($+5.75$ points; 95\% paired interval $[3.84,7.72]$), while GCD falls from $8.76^\circ$ to $7.83^\circ$. Multi-turn F5 rises from 604 to 610 ($+0.84$ points; $[-0.70,2.39]$), and GCD falls from $9.61^\circ$ to $8.48^\circ$ with four more valid answers. Multi-turn Hit remains near 91\% ($91.44\rightarrow91.16\%$).

\subsection{SSR training controls}
\label{subsec:ssr_results}
\input{Section/current_ssr_table}
Table~\ref{tab:current_ssr} shows substantial S360 gains from the SSR training recipe in both inference modes. Under CE-only supervision, S360 F5 rises from 70.41\% to 76.86\% for single-turn and from 71.39\% to 83.31\% for multi-turn training, gains of 6.45 and 11.92 percentage points. All-question GCD falls from $18.63^\circ$ to $8.18^\circ$ and from $20.88^\circ$ to $8.29^\circ$, respectively. SSR therefore improves both fixed-angle success and aggregate localization error, with a larger F5 gain for the multi-turn model. These results support SSR as an effective training augmentation for direction-balanced localization.

On APRS256, CE-only transfer effects are smaller and depend on the inference mode. Single-turn F5 increases from 66.41\% to 68.75\%, while GCD changes from $8.77^\circ$ to $9.37^\circ$. Multi-turn F5 decreases from 63.67\% to 62.50\% (three questions), with nearly unchanged GCD ($31.88^\circ$ versus $31.86^\circ$). The combined SSR+SDR recipe achieves the highest APRS256 F5 among these configurations in both modes (70.70/66.80\%) and lowers multi-turn GCD to $19.89^\circ$. These results highlight SSR's strong gains on direction-balanced S360 and the value of combining it with geometric supervision for transfer. Original-pool controls retain ordinary yaw augmentation; Table~\ref{tab:current_ssr} compares the full SSR recipe, including its additional training exposure. Further SSR--SDR ablation analyses are provided in the supplement.

\subsection{Spherical geometry under fixed SSR/SDR}
\label{subsec:sphere_results}
\input{Section/current_sphere_table}
Table~\ref{tab:current_sphere} fixes SSR/SDR and matches the capacity of the three added geometry branches. All variants retain native RoPE/MRoPE. TapPlane uses $(\theta,\phi,0)$, favoring proximity on the angular grid but treating opposite seam edges as distant. TapCylinder uses $(\cos\theta,\sin\theta,\phi)$, making longitude periodic while retaining latitude-independent horizontal distances. TapSphere uses unit rays $\mathbf d(\theta,\phi)$, capturing both seam continuity and the convergence of longitudes toward the poles. Native has no added geometric bias.

Each added branch applies a normalized negative squared-distance bias to the content-dependent Q/K logits before softmax (Eq.~\ref{eq:tapsphere_attention}).\par
\noindent With positive learned strengths, geometrically nearby patch pairs receive a smaller penalty than distant pairs. Thus the variants change which spatial relationships attention favors; the final weights still depend on visual content and the learned bias strength. The prior does not impose a fixed attention map.

TapSphere gains 3.12/7.03 F5 points over Native on APRS256 in single-/multi-turn inference and 4.91/1.12 on S360. On multi-turn APRS256, TapPlane/TapCylinder instead lose 7.03/7.42 points relative to Native. TapSphere therefore improves on Native in all four reported settings and leads both APRS256 comparisons, while TapCylinder remains slightly stronger on single-turn S360. These results favor spherical distance as an incremental prior beyond added capacity alone, particularly for transfer. These gains suggest that TapSphere's modeling of seam continuity and polar geometry provides a prior better suited to panoramic search than the planar and cylindrical alternatives. Further analyses are provided in the supplement.

\subsection{Auxiliary direction supervision}
\label{subsec:sdr_results}
With TapSphere and SSR fixed, adding SDR to CE improves F5 in both modes on S360 and APRS256 (Table~\ref{tab:current_final_sdr}). The gains are 0.70/1.95 percentage points for single-turn inference and 1.40/4.30 points for multi-turn inference, respectively. On APRS256, multi-turn SDR also lowers all-question GCD from $31.86^\circ$ to $19.89^\circ$ and increases valid answers from 219 to 236. These results support SDR as a geometric complement to token-level supervision, with improved task completion in multi-turn transfer. The supplement provides further CE/SDR ablations, comparisons with differentiable coordinate supervision, and paired validity/precision analyses.

\section{Conclusion}
\label{sec:discussion}
SPIRAL-360 combines SSR, TapSphere, and SDR in separately trained single-turn and multi-turn models, supported by S360-Search. Experiments demonstrate state-of-the-art S360 localization among evaluated systems and competitive object localization on APRS256 and H$^*$Bench. Together, the framework and dataset support efficient panoramic grounding and further study of visual tool reasoning.

\paragraph{Limitations.}
\label{sec:limitations}
Our study focuses on static panoramas. Extending SPIRAL-360 to panoramic video requires temporal reasoning and consistent target localization across frames. Natural scenes with targets near the viewing poles remain sparsely represented in our data, so generalization to these cases needs further validation. Broader video datasets and natural polar annotations offer promising directions for future work.\label{sec:main_end}\par

%% file: Section/2_related_work.tex
\section{Related work}
\paragraph{Panoramic grounding and search.}
Touchdown combines navigation with spatial description resolution, while Refer360 grounds instructions through dynamic partial views~\citep{Chen_2019_Touchdown,Cirik_2020_ACL_Refer360}. Hstar and Imagining360 study active search from limited views~\citep{Yu_2025_Thinking360,Zhang_2026_Imagining360}. PanoSeeker uses spherical memory for referring segmentation, and PAP uses panoramic-to-local routing for affordance prediction~\citep{tang2026seek,Zhang_2026_PAP}. Dense360 provides dense grounding supervision, PanoWorld predicts spherical locations from full panoramas, and GroundingPI outputs quantized image points and boxes~\citep{Zhou_2025_Dense360,Wang_2026_PanoWorld,yu2026groundingpi}. SPIRAL-360 predicts global target directions from a fixed panorama with separately trained direct-answer and tool-assisted models. Across prior tasks, both initial observations and prediction targets vary. A full panorama supplies global context, whereas local-view search acquires evidence through a sequence of observations. In our fixed-center setting, a tool changes viewing direction and field of view within the same scene, while the answer remains a global spherical direction. This connects direct grounding with visual search at a fixed optical center.

\paragraph{Panoramic reasoning.}
OSR-Bench, ODI-Bench, OmniCoT and PCSR-Bench evaluate omnidirectional understanding and spatial reasoning~\citep{Dongfang_2025_OSRBench,Yang_2026_ODIBench,He_2026_OmniCoT,Chen_2026_PCSR}. 360-R1 and PanoEnv apply reinforcement learning to panoramic question answering; \emph{Think in 360} uses space-time evidence to supervise panoramic video reasoning~\citep{Zhang_2025_360R1,Lin_2026_PanoEnv,Lu_2026_Think360}. Our task fixes the viewing center, whereas navigation benchmarks such as REVERIE and GOAT require movement between locations~\citep{Qi_2020_CVPR_REVERIE,Khanna_2024_CVPR_GOAT}.

\paragraph{Spherical geometry and supervision.}
Spherical CNNs and SphereUFormer model spherical features, while PanoSwin incorporates spherical distance into attention~\citep{Cohen_2018_ICLR_SphericalCNNs,Benny_2025_CVPR_SphereUFormer,Ling_2023_CVPR_PanoSwin}. RoPE Rolling, ERP-RoPE, FishRoPE and SpheRoPE adapt rotary embeddings~\citep{Su_2021_RoFormer} for reconstruction, omnidirectional perception or panorama generation~\citep{Ren_2025_ICCV_PanoSplatt3R,Zhou_2025_Dense360,Ahuja_2026_FishRoPE,Hirschorn_2026_SpheRoPE}. TapSphere retains native RoPE and adds ERP ray-distance biases; PanoWorld instead cross-attends to spherical tokens~\citep{Wang_2026_PanoWorld}. For augmentation, Pano Stretch and PanDA transform panoramas for layout and depth learning, while PanoVGGT jointly rotates images, depth and poses~\citep{Sun_2019_CVPR_HorizonNet,Cao_2025_PanDA,Guo_2026_PanoVGGT}. SSR couples directional stratification with consistent reconstruction of tool observations and coordinate-dependent text. SDR supervises physical target rays through an auxiliary head, related to cosine-based spherical gaze regression~\citep{Li_2021_GazeFollow360}. These geometric targets differ from the semantic embedding directions in angular object supervision~\citep{Esmaeilkhani_2026_AngularSupervision}.

\paragraph{Visual tools and policy learning.}
ReAct couples language reasoning with actions; V* and CogCoM acquire evidence through visual search and image operations~\citep{Yao_2023_ReAct,Wu_2024_VStar,Qi_2025_CogCoM}. Reinforcement learning supports crop-and-zoom policies~\citep{Wang_2025_PixelReasoner,Zheng_2026_DeepEyes,Zhu_2025_ActiveO3}, image annotation and masking~\citep{Wu_2026_VToolR1}. NV-CoT predicts continuous region actions, while OPO assigns credit to observation actions~\citep{Zhao_2026_NVCoT,Wang_2026_OPO}. MED distinguishes intrinsic capability gains from tool-induced effects~\citep{Ma_2026_MED}. We use terminal direction rewards and compare separately trained direct-answer and tool-assisted models by localization, answer validity and cumulative cost.

%% file: Section/3_tapsphere.tex
\subsection{TapSphere: Spherical visual attention}
\label{subsec:tapsphere_method}
ERP image-space distances misrepresent spherical proximity at the seam and near the poles. TapSphere introduces an angular-proximity prior into visual attention, related to PanoSwin's spherical-distance bias~\cite{Ling_2023_CVPR_PanoSwin}. The placeholder \texttt{<panoramic\_image>} expands to dedicated \texttt{<panoramic\_image\_pad>} tokens, and a learned type embedding augments merged ERP features. Perspective views use ordinary image tokens. All controls share this routing.

Each raw ERP patch center defines a unit ray $\mathbf d_i=\mathbf d(\theta_i,\phi_i)$. At selected vision block $\ell$, head $h$ adds a geometric bias to its attention logits:
\begin{equation}
 s_{ij}^{\ell h}
 =\frac{\mathbf q_i^{\ell h\top}\mathbf k_j^{\ell h}}{\sqrt{d_h}}
 -\frac{\kappa_{\ell h}}{2\rho_I}
       \|\mathbf d_i-\mathbf d_j\|_2^2,
 \qquad \kappa_{\ell h}>0,
\label{eq:tapsphere_attention}
\end{equation}
where $d_h$ is the native head width and $\mathbf q,\mathbf k$ include native visual RoPE. Row-wise softmax weights aggregate native value vectors. Normalization uses $\rho_I$, the row-centered RMS of $K_{ij}=-\|\mathbf d_i-\mathbf d_j\|_2^2/2$ over the patch grid. Since $\|\mathbf d_i-\mathbf d_j\|_2^2=2(1-\cos\gamma_{ij})$, the bias depends on angular separation $\gamma_{ij}$: it links seam-adjacent rays and accounts for longitude convergence near the poles.

The two terms in Eq.~\ref{eq:tapsphere_attention} encode complementary information: learned Q/K compatibility depends on visual content, while the added term reflects spherical proximity. Patches on opposite sides of the ERP seam can therefore incur a small geometric penalty despite their distant image columns. The same horizontal image offset can also imply different angular separations at different latitudes. This prior guides attention without asserting that nearby patches depict the same object.

We use visual blocks 8, 16, and 24 (zero-based), with 48 learned positive head strengths initialized to 0.25. Appending five deterministic geometric Q/K features implements Eq.~\ref{eq:tapsphere_attention} without a dense distance matrix. Native visual RoPE and decoder MRoPE remain active; the bias affects only ERP attention in training and inference. The kernel is invariant to a common rotation of its rays, although the full model need not be. The supplement details the construction and normalization.

Native omits the bias. TapPlane and TapCylinder replace unit rays with $(\theta,\phi,0)$ and $(\cos\theta,\sin\theta,\phi)$, with the same layers, parameter counts, and kernel normalization. These controls compare geometric priors at fixed SSR and SDR.

%% file: Section/3_geometry_aware_training.tex
\subsection{Spherically stratified replay}
\label{subsec:geometry_training}
\label{subsec:ssr_method}
SSR broadens directional coverage through stratified replay and trajectory reconstruction. Six target strata---central-left/right, seam-left/right, and north/south polar---each contribute 469 replay trajectories. The 938 polar examples reorient targets to elevations $\pm75^\circ$; the 1,876 nonpolar examples retain their source panoramas and directions. These equal quotas balance the replay pool added to 21,884 original trajectories.

A rotation $R\in SO(3)$ jointly transforms the panorama, target ray $\mathbf d^*$, and tool viewing-axis ray $\mathbf c_t$:
\begin{equation}
I_R(\mathbf d)=I(R^{-1}\mathbf d),\quad
\mathbf d^{*,R}=R\mathbf d^*,\quad
\mathbf c_t^R=R\mathbf c_t.
\label{eq:ssr_rotation}
\end{equation}
For polar reorientation, we align the target-centered canonical frames:
$R=C(\mathbf d^{*,R})C(\mathbf d^*)^\top$, where $C$ contains right, up, and forward axes. This fixes residual roll, preserving local orientation in a target-centered perspective view up to resampling. Equation~\ref{eq:ssr_rotation} preserves target--viewing-axis angles, although off-target images may rotate.

Rotating only the initial ERP would leave later camera directions and terminal coordinates in the old frame. Reconstructing the complete trajectory instead ties each rendered observation to its transformed action and keeps the final target label consistent.

We resample the ERP with longitude wrapping and re-render tool views using canonical zero-roll cameras at their original FoVs. Action coordinates and terminal answers use the transformed frame; annotation-derived geometric templates replace reasoning text. Nonpolar replay uses the same reconstruction without changing directions; original trajectories remain intact. This keeps panoramas, actions, observations, and targets geometrically consistent, without guaranteeing unchanged target visibility in rectangular views or visually verified reasoning. The supplement details the strata and reconstruction.

\subsection{Auxiliary direction supervision}
\label{subsec:sdr_method}
SDR adds target-direction supervision before each reply. For reply $t$ in trajectory $i$, we extract $\mathbf h_{it}$ immediately before its supervised span, using the final decoder block's output before final normalization. A training-only affine head predicts
$\mathbf z_{it}=A\mathbf h_{it}+\mathbf b$ and
$\widehat{\mathbf d}_{it}=\mathbf z_{it}/\max(\|\mathbf z_{it}\|_2,10^{-6})$.
For $B$ trajectories, with $T_i$ replies in trajectory $i$, we minimize
\begin{equation}
\mathcal L_{\rm SDR}
=\frac{1}{B}\sum_{i=1}^{B}\frac{1}{T_i}\sum_{t=1}^{T_i}
 (1+\sin^2\phi_i^*)\!
 \left[1-\operatorname{clip}_{[-1,1]}
 (\widehat{\mathbf d}_{it}^{\top}\mathbf d_i^*)\right].
\label{eq:ccsar_loss}
\end{equation}
Here $\mathbf d_i^*$ and $\phi_i^*$ are the target direction and elevation after replay. The stabilized cosine penalty in Eq.~\ref{eq:ccsar_loss} measures directional agreement and upweights polar latitudes. Averaging over replies makes each trajectory's auxiliary weight independent of its reply count. Each anchor attends to earlier observations and replies, excluding the response being predicted.

For a tool-call reply, the auxiliary target remains the object's global direction, even when the requested camera points elsewhere. Across a trajectory, successive anchors therefore receive the same geometric target while conditioning on the evolving observation history. This separates target localization from camera-action prediction: the direction head supplies a geometric training signal, while cross-entropy continues to supervise the actual tool arguments and terminal coordinate tokens.

SDR gradients update only the direction head for the first ten updates, then also reach the backbone. Cross-entropy trains the full model throughout. SDR supervises contextual direction representations, not the generated coordinate digits. At inference, we discard the head and evaluate the language-generated answer. The supplement details normalization and gradient flow.

%% file: Section/3_supervised_modes.tex
\subsection{Training single-turn and multi-turn models}
\label{subsec:training}
We derive both training datasets from the same 24,698 trajectories, preserving their order and SSR ratio. Each trajectory contains an initial panorama, query, assistant responses, and rendered tool observations.

\paragraph{Single-turn supervision.}
We retain the first panorama and original question, replace the tool instruction with a direct-answer instruction, and copy the final answer verbatim. Later observations, tool actions, and teacher reasoning are omitted. Cross-entropy supervises the answer and message terminator; one SDR anchor precedes the answer.

\paragraph{Multi-turn supervision.}
We serialize the full trajectory as one causal sequence. Cross-entropy supervises assistant reasoning, tool-call arguments, final answers, and terminators. User messages, tool observations, and visual tokens provide context and are masked from the language loss. Each assistant turn has an SDR anchor before its supervised span. Training conditions on recorded histories; evaluation follows generated actions and rendered observations.

For either mode, $\mathcal A_i$ indexes supervised assistant tokens in example $i$. We optimize
\begin{align}
\mathcal L_{\rm CE}
&=-\frac{1}{B}\sum_{i=1}^{B}\frac{1}{|\mathcal A_i|}
  \sum_{j\in\mathcal A_i}
  \log p_\vartheta(y_{ij}\mid x_{i,<j}),\notag\\
\mathcal L_{\rm SFT}
&=\mathcal L_{\rm CE}+\lambda_{\rm SDR}\mathcal L_{\rm SDR},
\label{eq:ccsar_total}
\end{align}
where $x_{i,<j}$ is the preceding multimodal context. Averaging first over each trajectory's supervised tokens and then over the batch gives trajectories equal weight. In Eq.~\ref{eq:ccsar_total}, $\lambda_{\rm SDR}=0.02$, or zero for matched CE-only ablations.

Both models independently start from Qwen3-VL-8B-Instruct~\cite{Qwen3-VL}, sharing the base checkpoint, source examples, initial-image budgets, and optimizer settings. Each receives one full-parameter SFT epoch, with a global batch size of eight and 3,088 updates. Multi-turn training uses more visual and supervised tokens because it retains full trajectories. The supplement details this protocol and the separate replay continuation study.

\paragraph{Terminal-reward post-training.}
After SFT, GRPO~\cite{Shao2024DeepSeekMathPT} samples four completions per training question for each mode. Valid terminal answers receive a fixed-$5^\circ$ success reward plus a continuous angular reward; invalid outputs are penalized. We update only language-model LoRA~\cite{hu2022lora} parameters, freeze visual geometry, and select checkpoints on DEV319. GRPO evaluation is separate from the SFT component controls. The supplement specifies the reward and training budget.

%% file: Section/current_main_results_table.tex
\begin{table*}[!t]
\centering
\caption{\textbf{System comparison on S360.} Results on 713 direction-balanced questions; metrics are defined in Section~\ref{subsec:experiment_setup}. The top three distinct values are \colorbox{red!15}{first}, \colorbox{green!15}{second}, and \colorbox{blue!15}{third}; bold denotes the best, including ties. Prompts and observation budgets differ (see supplement). $\dagger$: provider-internal reasoning. $\ddagger$: tool-enabled; HVS/Imagining360 start from a local view, ours from full ERP. $\star$: GroundingPI inputs have 710 identical file hashes and three further images with identical decoded pixels.}
\label{tab:main_experiment}
\label{tab:current_main_results}
\small
\renewcommand{\arraystretch}{1.05}
\setlength{\tabcolsep}{2.5pt}
\begin{tabular*}{\textwidth}{@{\extracolsep{\fill}}lrrrrrrrr@{}}
\toprule
Method & Hit & GCD & \multicolumn{6}{c}{Direction-wise Hit (\%)$\uparrow$} \\
\cmidrule(lr){4-9}
& (\%)$\uparrow$ & ($^\circ$)$\downarrow$ & Front & Back & Left & Right & Up & Down \\
Samples & 713 & 713 & 119 & 119 & 119 & 119 & 119 & 118 \\
\midrule
\multicolumn{9}{l}{\textbf{Open-source models (archived)}} \\
InternVL3.5-4B & 2.52 & 102.34 & 0.00 & 5.88 & 2.52 & 4.20 & 0.84 & 1.69 \\
InternVL3.5-8B & 5.33 & 78.46 & 11.76 & 4.20 & 5.04 & 6.72 & 1.68 & 2.54 \\
Qwen3-VL-4B-Instruct & 7.57 & 72.73 & 20.17 & 11.76 & 8.40 & 5.04 & 0.00 & 0.00 \\
Qwen3-VL-8B-Instruct & 5.61 & 78.34 & 2.52 & 17.65 & 6.72 & 5.88 & 0.00 & 0.85 \\
Qwen3.5-4B & 9.26 & 53.96 & 10.08 & 13.45 & 11.76 & 10.08 & 7.56 & 2.54 \\
Qwen3.5-9B & 16.41 & 40.30 & 27.73 & 19.33 & 14.29 & 17.65 & 10.08 & 9.32 \\
\midrule
\multicolumn{9}{l}{\textbf{Proprietary models (archived)}} \\
GPT-4o & 9.12 & 52.15 & 9.24 & 0.00 & 14.29 & 25.21 & 3.36 & 2.54 \\
Gemini 2.5 Pro$^{\dagger}$ & 17.95 & 34.97 & 10.08 & 15.13 & 26.05 & 21.85 & 21.85 & 12.71 \\
Claude Sonnet 5 & 17.67 & 39.90 & 14.29 & 47.06 & 6.72 & 17.65 & 11.76 & 8.47 \\
GPT-5.5 & 31.84 & 25.43 & 34.45 & 48.74 & 25.21 & 51.26 & 13.45 & 17.80 \\
Gemini 3.1 Pro Preview$^{\dagger}$ & 48.81 & 15.88 & 67.23 & 57.14 & 42.02 & 35.29 & \cellcolor{blue!15}43.70 & 47.46 \\
GPT-5.6 Sol & 71.53 & \cellcolor{blue!15}15.06 & \cellcolor{blue!15}89.92 & 81.51 & 89.92 & \cellcolor{blue!15}89.08 & \cellcolor{blue!15}43.70 & 34.75 \\
\midrule
\multicolumn{9}{l}{\textbf{Panoramic comparison methods}} \\
PanoWorld-9B (public) & 69.14 & 17.56 & \cellcolor{green!15}95.80 & \cellcolor{green!15}96.64 & 92.44 & \cellcolor{red!15}\textbf{98.32} & 16.81 & 14.41 \\
Imagining360 (public)$^{\ddagger}$ & 34.64 & 47.46 & 57.14 & 43.70 & 47.06 & 47.90 & 1.68 & 10.17 \\
HVS-3B (public)$^{\ddagger}$ & 14.73 & 73.05 & 13.45 & 27.73 & 26.89 & 20.17 & 0.00 & 0.00 \\
GroundingPI (public)$^{\star}$ & \cellcolor{blue!15}82.75 & 16.25 & \cellcolor{green!15}95.80 & \cellcolor{blue!15}94.12 & \cellcolor{red!15}\textbf{98.32} & \cellcolor{red!15}\textbf{98.32} & \cellcolor{green!15}48.74 & \cellcolor{blue!15}61.02 \\
\midrule
\multicolumn{9}{l}{\textbf{Our method}} \\
SPIRAL-360-8B (single-turn GRPO) & \cellcolor{green!15}89.62 & \cellcolor{red!15}\textbf{7.83} & \cellcolor{red!15}\textbf{96.64} & 91.60 & \cellcolor{green!15}97.48 & \cellcolor{red!15}\textbf{98.32} & \cellcolor{red!15}\textbf{80.67} & \cellcolor{green!15}72.88 \\
SPIRAL-360-8B (multi-turn GRPO)$^{\ddagger}$ & \cellcolor{red!15}\textbf{91.16} & \cellcolor{green!15}8.48 & \cellcolor{red!15}\textbf{96.64} & \cellcolor{red!15}\textbf{97.48} & \cellcolor{blue!15}96.64 & \cellcolor{green!15}97.48 & \cellcolor{red!15}\textbf{80.67} & \cellcolor{red!15}\textbf{77.97} \\
\bottomrule
\end{tabular*}
\end{table*}

%% file: Section/current_external_results_table.tex
\begin{table}[!htbp]
\centering
\caption{\textbf{H$^*$Bench search success (\%).} HOS/HPS denote object/path search; $\dagger$ marks H$^*$-specific training.}
\label{tab:current_external_results}
\label{tab:current_hstar_results}
\small
\renewcommand{\arraystretch}{1.05}
\setlength{\tabcolsep}{3pt}
\begin{tabular*}{\columnwidth}{@{\extracolsep{\fill}}lrr@{}}
\toprule
Method & HOS$\uparrow$ & HPS$\uparrow$ \\
\midrule
\multicolumn{3}{l}{\emph{Local views: Hstar~\cite{Yu_2025_Thinking360}}} \\
InternVL3.5-4B & 3.08 & 4.81 \\
InternVL3.5-8B & 6.38 & 7.25 \\
Qwen2.5-VL-3B & 14.83 & 6.44 \\
Qwen2.5-VL-7B & 11.38 & 6.31 \\
Gemma-3-4B & 17.13 & 14.44 \\
Gemma-3-12B & 10.21 & 14.50 \\
Kimi-VL-A3B & 4.92 & 4.32 \\
GPT-4o & 19.75 & 23.69 \\
Gemini 2.5 Pro & 31.96 & 33.00 \\
HVS-3B (SFT)$^{\dagger}$ & 40.83 & 23.00 \\
HVS-3B (SFT+RL)$^{\dagger}$ & 47.38 & 24.94 \\
\midrule
\multicolumn{3}{l}{\emph{Local views: Imagining360~\cite{Zhang_2026_Imagining360}}} \\
HVS-3B (RL) + Imaginator-8B$^{\dagger}$ & 62.75 & 39.38 \\
Gemini 2.5 Pro + Imaginator-8B$^{\dagger}$ & 66.10 & 43.10 \\
\midrule
\multicolumn{3}{l}{\emph{Full ERP: PanoWorld~\cite{Wang_2026_PanoWorld}}} \\
PanoWorld-9B (zero-shot) & 61.8 & 47.5 \\
PanoWorld-9B + H$^*$ SFT$^{\dagger}$ & 73.1 & 64.2 \\
\midrule
\multicolumn{3}{l}{\emph{Full ERP: SPIRAL-360}} \\
SPIRAL-360-8B (single GRPO) & 68.33 & 25.25 \\
SPIRAL-360-8B (multi GRPO) & 66.17 & 24.25 \\
\bottomrule
\end{tabular*}
\end{table}

%% file: Section/current_ssr_table.tex
\begin{table}[!htbp]
\centering
\caption{\textbf{SSR and SDR with full-parameter SFT.} Detailed experimental settings and analyses are provided in the supplement.}
\label{tab:current_ssr}
\label{tab:current_final_sdr}
\small
\setlength{\tabcolsep}{3pt}
\begin{tabular*}{\columnwidth}{@{\extracolsep{\fill}}lrrrr@{}}
\toprule
 & \multicolumn{2}{c}{S360} & \multicolumn{2}{c}{APRS256} \\
\cmidrule(lr){2-3}\cmidrule(lr){4-5}
Training configuration & F5$\uparrow$ & GCD$\downarrow$ & F5$\uparrow$ & GCD$\downarrow$ \\
\midrule
\multicolumn{5}{l}{\emph{Single-turn TapSphere SFT}} \\
Original + CE-only & 70.41 & 18.63 & 66.41 & \textbf{8.77} \\
SSR + CE-only & 76.86 & \textbf{8.18} & 68.75 & 9.37 \\
SSR + CE+SDR & \textbf{77.56} & 8.76 & \textbf{70.70} & 9.13 \\
\midrule
\multicolumn{5}{l}{\emph{Multi-turn TapSphere SFT}} \\
Original + CE-only & 71.39 & 20.88 & 63.67 & 31.88 \\
SSR + CE-only & 83.31 & \textbf{8.29} & 62.50 & 31.86 \\
SSR + CE+SDR & \textbf{84.71} & 9.61 & \textbf{66.80} & \textbf{19.89} \\
\bottomrule
\end{tabular*}
\end{table}

%% file: Section/current_sphere_table.tex
\begin{table}[!htbp]
\centering
\caption{\textbf{Geometry controls at fixed SSR/SDR.} F5 (\%) on S360 and APRS256. All models are evaluated after SFT. Multi-turn APRS256 retains its earlier frozen evaluation; Table~\ref{tab:current_final_sdr} uses a later paired evaluation of the same TapSphere checkpoint (see supplement).}
\label{tab:current_sphere}
\small
\setlength{\tabcolsep}{3pt}
\begin{tabular*}{\columnwidth}{@{\extracolsep{\fill}}lrrrr@{}}
\toprule
& \multicolumn{2}{c}{Single-turn} & \multicolumn{2}{c}{Multi-turn} \\
\cmidrule(lr){2-3}\cmidrule(lr){4-5}
Encoding & S360 & APRS256 & S360 & APRS256 \\
\midrule
Native & 72.65 & 67.58 & 83.59 & 60.16 \\
TapPlane & 77.14 & 69.14 & 83.17 & 53.12 \\
TapCylinder & \textbf{77.98} & 66.41 & 83.31 & 52.73 \\
TapSphere & 77.56 & \textbf{70.70} & \textbf{84.71} & \textbf{67.19} \\
\bottomrule
\end{tabular*}
\end{table}

%% file: Section/ai_use_statement.tex
\section*{LLM Usage Statement}
We used OpenAI ChatGPT/GPT to obtain reviewer-style feedback on substantial portions of the manuscript and OpenAI Codex (including GPT-6.1 Sol) to draft, restructure, translate, and edit manuscript and supplementary text; find related work, generate summaries and draft citation records, and check citations; and generate implementation and data-processing code, experiment protocols, and analyses. Codex also helped debug code and launch, monitor, and resume training and evaluation runs. Authors reviewed, edited, or approved the generated material. Codex also helped formulate and refine hypotheses, derive and check geometric relationships and training objectives, and edit conclusions. It generated editable layouts and vector annotations while refining figures to author specifications. The teaser and pipeline use source panoramas and geometrically rendered perspective views.

GPT-4o mini generated candidate target descriptions and helped check uniqueness on real panoramas; Qwen3-VL supported synthetic-example screening. Earlier staged data development also used offline teacher- and student-generated dialogue prefixes. The current full-data comparisons use complete geometry-guided trajectories. The supplement details data construction and training. Checks included executing code and tests, checking coordinate and data consistency, comparing reported scores with saved predictions, and consulting primary sources. The authors take full responsibility for all submitted content, including the paper's claims, results, and reproducibility.